%% file: main.tex
\documentclass[10pt,twocolumn,letterpaper]{article}

\usepackage{cvpr}              %
\usepackage{arydshln}
\usepackage{makecell}
\usepackage{booktabs}
\usepackage{multirow}
\usepackage{array}
\usepackage{csquotes}
\usepackage{subcaption}
\usepackage{comment}
\usepackage[table]{xcolor}  %
\input{preamble}

\definecolor{cvprblue}{rgb}{0.21,0.49,0.74}
\usepackage[pagebackref,breaklinks,colorlinks,allcolors=cvprblue]{hyperref}

\def\paperID{} %
\def\confName{CVPR}
\def\confYear{2026}

\title{See It, Say It, Sorted: An Iterative Training-Free Framework for Visually-Grounded Multimodal Reasoning in LVLMs}

\author{
Yongchang Zhang\textsuperscript{1,3\textasteriskcentered}, Oliver Ma\textsuperscript{2\textasteriskcentered}, Tianyi Liu\textsuperscript{1\textdagger}, Guangquan Zhou\textsuperscript{1\textdagger}, Yang Chen\textsuperscript{1\textdagger} \\
\textsuperscript{1}Southeast University \quad
\textsuperscript{2}University of Oxford \quad
\textsuperscript{3}AIIA, Ministry of Education, China \quad \\
{\tt\small \textbf{yongchangzhang2005@gmail.com \quad chenyang.list@seu.edu.cn}} \\
{\tt\small \url{https://github.com/uuuuZYC/See-It-Say-It-Sorted}}
\vspace{-7pt}
}

\begin{document}
\maketitle

\begingroup
\renewcommand{\thefootnote}{}%
\footnotetext{%
\textsuperscript{\textasteriskcentered}Equal contribution \quad
\textsuperscript{\textdagger}Corresponding author%
}%
\endgroup

\input{sec/0_abstract}

\input{sec/1_introduction}

\input{sec/3_methodology}

\input{sec/4_experiment}

\input{sec/5_conclusion}
\section*{Acknowledgements}
This work was supported in part by the National Natural Science Foundation of China under Grant T2225025 and 62371121, in part by the National Key Research and Development Program of China under Grant 2024YFF1206700, and in part by the Fundamental Research Funds for the Central Universities. AIIA refers to the Key Laboratory of New Generation Artificial Intelligence Technology and Its Interdisciplinary Applications (Southeast University).

{
    \small
    \bibliographystyle{ieeenat_fullname}
    \bibliography{main}
}

\end{document}

%% file: preamble.tex
\usepackage{arydshln}

%% file: sec/0_abstract.tex
\begin{abstract}
Recent large vision-language models (LVLMs) have demonstrated impressive reasoning ability by generating long chain-of-thought (CoT) responses. However, CoT reasoning in multimodal contexts is highly vulnerable to visual hallucination propagation: once an intermediate reasoning step becomes inconsistent with the visual evidence, subsequent steps—even if logically valid—can still lead to incorrect final answers.
Existing solutions attempt to mitigate this issue by training models to “think with images” via reinforcement learning (RL). While effective, these methods are costly, model-specific, and difficult to generalize across architectures.
Differently, we present a lightweight method that bypasses RL training and provides an iterative, training-free, plug-and-play framework for visually-grounded multimodal reasoning. Our key idea is to supervise each reasoning step at test time with visual evidence, ensuring that every decoded token is justified by corresponding visual cues. Concretely, we construct a textual visual-evidence pool that guides the model’s reasoning generation. When existing evidence is insufficient, a visual decider module dynamically extracts additional relevant evidence from the image based on the ongoing reasoning context, expanding the pool until the model achieves sufficient visual certainty to terminate reasoning and produce the final answer.
Extensive experiments on multiple LVLM backbones and benchmarks demonstrate the effectiveness of our approach. Our method achieves 16.5\%–29.5\% improvements on TreeBench and 13.7\% RH-AUC gains on RH-Bench, substantially reducing hallucination rates while improving reasoning accuracy without additional training.
\end{abstract}

%% file: sec/1_introduction.tex
\begin{figure}[t]
\centering
\captionsetup{skip=4pt}
\includegraphics[width=1.0\columnwidth]{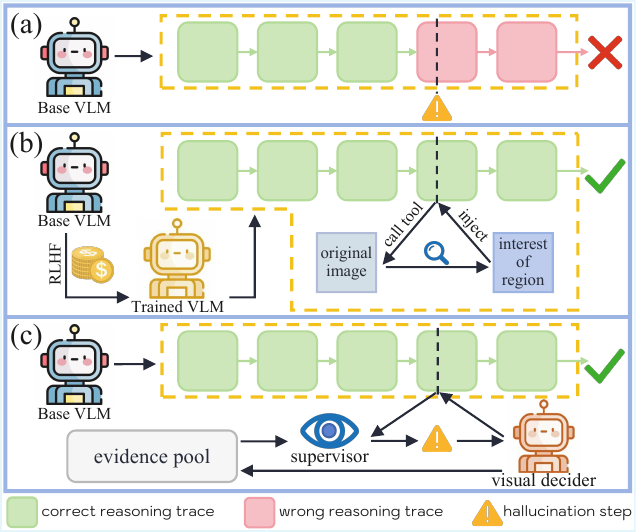}
\caption{
\textbf{Reasoning pattern comparison.} 
(a) Greedy decoding: the base VLM selects the top-1 token at each step; any hallucination in an intermediate step propagates to an incorrect final answer.
(b) RLHF-based “think-with-images”: the model learns when to call tools to zoom or crop the image and re-inject cropped regions into the reasoning context—effective but costly and model-specific.
(c) Ours: a lightweight, training-free, model-agnostic framework. A supervisor maintains a dynamic visual-evidence pool to detect and correct hallucination steps. When uncertainty arises, it invokes a visual decider to extract new evidence, enabling visually grounded reasoning throughout the chain.
}
\vspace{-12pt}
\label{fig1}
\end{figure}

\section{Introduction}
\label{sec:introduction}

Large vision--language models (LVLMs) now generate long chain-of-thought (CoT) explanations and solve diverse multimodal reasoning tasks~\cite{team2025kimi,team2025kimivl,guo2025seed1,shao2024visual, sun2024visual}. Yet, the very ability to ``think more'' often coincides with ``seeing less''.~\cite{liu2025thinkingseeingassessingamplified} During inference-time decoding, a model must balance three competing contexts: the image, a growing textual context, and instruction tokens. As the context lengthens, subtle but decisive visual cues are easily dominated by language priors. 
Even a single token that departs from the visual evidence can steer the remaining chain of thought toward a fluent but visually inconsistent trajectory (\cref{fig1} (a)).
This reasoning–perception drift originates at decoding time—not because the model lacks visual understanding, but because long-horizon token generation gradually amplifies language priors over visual grounding~\cite{wei2022chain,liu2023visual,li2023blip,rohrbach2018object}.

A prevailing solution is to explicitly train models to “think with images” by learning when and where to zoom or crop during CoT generation (\cref{fig1}(b)).
These RL- or preference-optimization pipelines train a policy to call visual tools, inspect regions, and re-inject pixels into the reasoning context.
While effective, they rely on curated data, reward design, heavy computation, and tight coupling to specific backbones.
The learned policy intertwines spatial exploration with the CoT, repeatedly encoding cropped views and incurring latency.
As model scale increases, such visually grounded training becomes prohibitively expensive, limiting accessibility for broader use.

To address these limitations, we pursue a different principle: rather than learning when to look at training time, we supervise each reasoning step with visual evidence at test time (\cref{fig1} (c)).
We introduce an iterative, training-free, plug-and-play framework that treats decoding as a sequence of evidence-justified token selections.
The system maintains a textual evidence pool that operates alongside the base LVLM.
At each step, the LVLM proposes a compact top-k set of candidate tokens from its local probability distribution.
A lightweight supervisor computes an evidence-induced preference over these candidates and negotiates a reweighted distribution with the base probabilities, preserving confident behavior while reallocating probability mass toward tokens consistent with accumulated evidence.
When residual uncertainty remains, a visual decider inspects the image under the current reasoning context and generates a concise micro-observation in natural language.
This observation is appended to the evidence pool and reused in all subsequent reasoning steps.

This design exhibits three key properties that directly mitigate the limitations of RL-based pipelines such as PixelReasoner~\cite{su2025pixel} and DeepEyes~\cite{zheng2025deepeyes}.
First, it is training-free and inherently transferable: the framework wraps around a frozen LVLM with a lightweight decider, requiring no task-specific finetuning or policy optimization.
Second, it is cost-aware by construction. A simple uncertainty test on the negotiated distribution determines whether to invoke the decider, ensuring that additional visual computation occurs only when most likely to prevent a hallucination.
Third, it represents evidence in text rather than pixels, enabling subsequent tokens to directly reference prior micro-observations without repeatedly re-encoding image crops. This textual form makes the framework easier to deploy at inference time and substantially reduces computational overhead compared with pixel-level reasoning.

Empirically, the framework improves both grounding and end-task accuracy across backbones and benchmarks while keeping overhead modest. Because evidence is accumulated on demand, fine-grained cues can be reused downstream to stabilize the remainder of the chain. Qualitative analyses show that many previously cascading failures reduce to one or two decisive micro-observations that the decider contributes precisely at uncertain steps; 
Quantitatively, we observe a clear accuracy–latency trade-off as the uncertainty threshold varies, allowing flexible adaptation to different deployment budgets.

The main contributions are summarized as follows:
\begin{itemize}
    \item We present a training-free, plug-and-play decoding framework that supervises token selection with a growing textual evidence pool and negotiates next-token probabilities with the base model rather than relying on learned tool-calling policies.
    \item An uncertainty-triggered visual decider emits concise, reusable micro-evidence only when necessary, yielding strong cost–accuracy trade-offs. 
    \item The approach transfers across LVLM backbones and consistently reduces hallucination while obviously improving task accuracy on a wide range of benchmarks.
\end{itemize}

\begin{figure*}[t]
\centering
\captionsetup{skip=4pt}
\includegraphics[width=2.1\columnwidth]{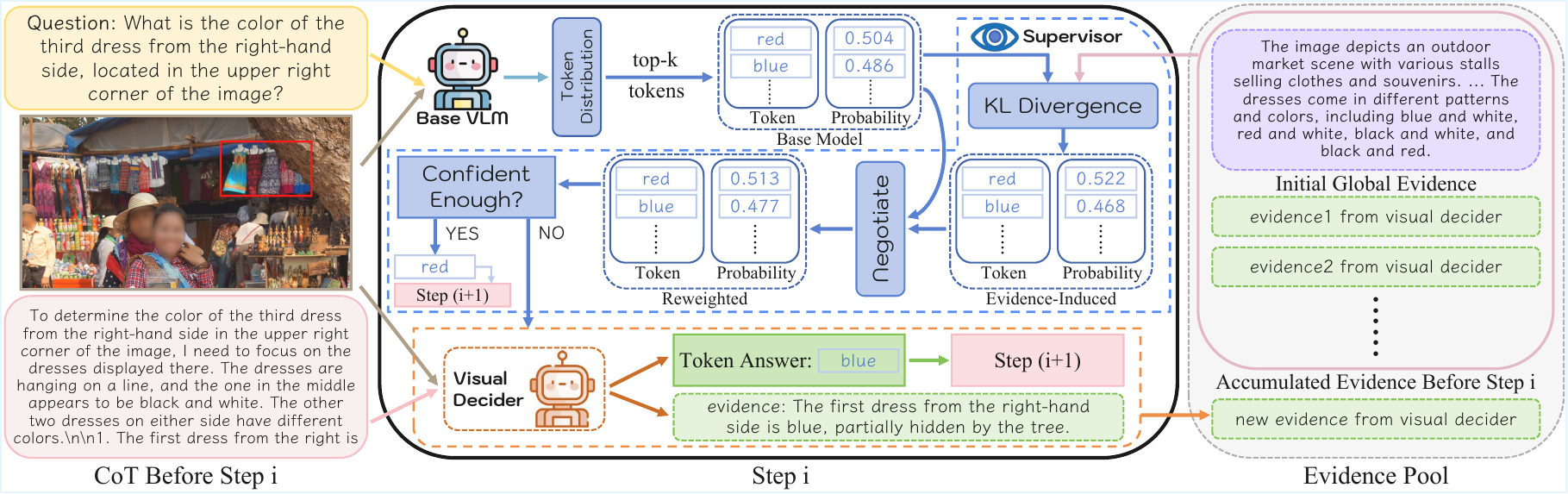}
\caption{Overview of evidence-constrained reweighting decoding (ECRD) at decoding step $i$. The base VLM emits a top-k candidate set; the supervisor builds an evidence-induced distribution from the current evidence pool and negotiates with the base probabilities to reweight candidates. If confidence remains low, the visual decider reads the image with the current prefix, commits a token, and adds a short textual evidence for later steps.}
\vspace{-8pt}
\label{fig2}
\end{figure*}

%% file: sec/3_methodology.tex
\section{Methodology}
\label{sec:methodology}

\subsection{Visual Description Grounded Decoding}

Let a VLM define, at decoding step \(i\), a next-token distribution:
\vspace{-8pt}
\begin{equation}
p_i(w) \;=\; p_{\text{VLM}}(x_i{=}w \mid x_{<i}), \qquad w\in\mathcal V .
\vspace{-4pt}
\end{equation}
Visual description grounded decoding (VDGD) first asks the model to produce one global textual description \(d=(d_1,\dots,d_L)\) of image and then constrains decoding by knee truncation to select $\text{top-}k$ plausible tokens and a description-based KL preference. If \(p_{(1)}\!\ge\!p_{(2)}\!\ge\!\cdots\) are sorted probabilities of \(p_i\), the knee index can be defined as follows:
\vspace{-4pt}
\begin{equation}
k^\star = \arg\max_k\big(p_{(k)}-p_{(k+1)}\big).
\vspace{-4pt}
\end{equation}
And then we can define the candidate set:
\vspace{-4pt}
\begin{equation}
\mathcal C_i=\text{Top-}k^\star\!\big(p_i\big).
\vspace{-4pt}
\end{equation}
For a candidate token \(w\in\mathcal C_i\) and a prefix of description length \(j\) ($1 \le j \le L$), VDGD measures the deviation:
\vspace{-4pt}
\begin{equation}
\mathrm{KL}\!\left(\mathrm{onehot}(w)\,\big\|\,p_{\text{VLM}}(\cdot\mid d_{<j})\right)
= -\log p_{\text{VLM}}(w\mid d_{<j}),
\end{equation}
then takes a minimum value over \(j\) (the best prefix) and replaces the base logits with these values before softmax. While effective in its one-shot setting, fixed logit replacement tied to a single static caption is less self-adaptive, and it does not preserve the base model’s calibrated confidence on already-confident steps.

\subsection{Distribution Supervisor}

As shown in \cref{fig2}, The distribution supervisor keeps the spirit of description-guided preference yet makes two principled changes: (a) we convert VDGD’s min-over-prefix KL into a mean-over-prefix probability that is further averaged across multiple evidences, and (b) we mix the evidence-induced distribution with the base model instead of overwriting logits.

\textbf{From VDGD’s KL to ECRD’s evidence score.}
Let \(\mathcal{E}_i=(e_1,\dots,e_L)\) be a piece of evidence sentence. 
VDGD’s min-over-prefix KL is well suited to the static captioning regime where the most supportive partial description is selected. In our dynamic regime, where evidence accrues incrementally over steps and will be mixed with the base distribution, thus, we need an aggregator that rewards sustained support across the sentence rather than one sharp peak without robustness, and compiles smoothly across multiple evidence. We therefore replace the min with a mean-over-prefix probability:
\begin{equation}
\label{eq5}
q_\mathcal{E}(w) \;=\; \frac{1}{L}\sum_{j=1}^{L} p_{\text{VLM}}(w \mid e_{<j}).
\end{equation}
Given a pool \(E_i=\{\mathcal{E}_1,\dots,\mathcal{E}_N\}\) of evidence available at step \(i\), we average their supports:
\begin{equation}
\label{eq6}
S_i(w) \;=\; -\log\!\Big(\tfrac{1}{N}\sum_{\mathcal{E}} q_\mathcal{E}(w)\Big),
\end{equation}
and restrict to \(\mathcal C_i\) to obtain the evidence-induced distribution:
\begin{equation}
\label{eq7}
r_i(w)=
\begin{cases}
\displaystyle \dfrac{\exp\{-S_i(w)\}}{\sum_{u\in\mathcal C_i}\exp\{-S_i(u)\}}, & w\in\mathcal C_i,\\[8pt]
0, & \text{otherwise}.
\end{cases}
\end{equation}

\textbf{Negotiated reweighting.}
Let \(p_i\) be the base distribution. We form a mass-matched version \(\tilde r_i\) , the rationale is as follows: \(p_i\) and the evidence-induced distribution \(r_i\) are not on the same scale. \(r_i\) is normalized only over the candidate set \(\mathcal C_i\) (and zero elsewhere), whereas \(p_i\) spreads probability over the full vocabulary. We therefore perform a proportional rescaling of \(r_i\) within \(\mathcal C_i\) so that the total mass assigned to \(\mathcal C_i\) matches that of \(p_i\). The resulting scaled distribution is denoted \(\tilde r_i\). Concretely,
\begin{equation}
\tilde r_i(w) \;=\;
\begin{cases}
\displaystyle r_i(w)\cdot \frac{\sum_{u\in\mathcal C_i} p_i(u)}{\sum_{u\in\mathcal C_i} r_i(u)}, & w\in\mathcal C_i,\\[6pt]
0, & \text{otherwise},
\end{cases}
\end{equation}
which ensures that we only reallocate the probability mass within \(\mathcal C_i\) without varying its total amount:
\begin{equation}
\sum_{w\in\mathcal C_i}\tilde r_i(w)=\sum_{w\in\mathcal C_i}p_i(w).
\end{equation}

and then let the base distribution and the evidence-induced distribution negotiate a mixture:
\begin{equation}
\label{eq10}
p_i^{\text{mix}}(w) =
\begin{cases}
\alpha_i\,p_i(w) + (1-\alpha_i)\,\tilde r_i(w), & w\in\mathcal C_i,\\[4pt]
\alpha_i\,p_i(w), & w\notin\mathcal C_i.
\end{cases}
\end{equation}
The adaptive weight is chosen without hyper-parameters as:
\begin{equation}
\alpha_i \;=\; p_{(1)},
\end{equation}
the top probability of the base model before mixing. This choice aligns with empirical statistics reported by~\cite{ghosh2024visual}: hallucination steps tend to have larger knee-selected \(k^\star\) and diffuse local distributions with small variance. However, non-hallucination steps are sharply peaked with average \(k^\star\!\approx\!1\). Consequently, when \(p_{(1)}\) is large (confident step), we keep the base distribution predominates; when \(p_{(1)}\) is small (hallucination-critical moments), evidence receives more weight. It reflects a general intervention principle: when the base distribution is sharp, evidence should act as a light prior; when it is diffuse, evidence should carry more weight. This makes mixture responsive to local uncertainty while preserving the behavior of base model in easy steps.

\textbf{When to acquire new evidence.}
After \cref{eq10}, we inspect the negotiated margin:
\begin{equation}
\label{eq12}
\Delta_i \;=\; p^{\text{mix}}_{(1)} - p^{\text{mix}}_{(2)} .
\end{equation}
If \(k^\star>1\) and \(\Delta_i \le \delta\) ($\delta$ is a hyperparameter), we consider step \(i\) a probable trigger for hallucination and acquire new evidence. Otherwise, we directly select \(p^{\text{mix}}_{(1)}\).

\subsection{Dynamic evidence pool and the visual decider}

\textbf{Invocation and interface.}
When the trigger in \cref{eq12} fires, we call a lightweight visual decider instantiated by GRIT~\cite{fan2025grit}, which built on Qwen2.5-VL-3B~\cite{bai2025qwen25vl}, except for the basic answer, it can optionally output the coordinates of one or more image regions that the model needs to refer to during its reasoning process when generating an answer. GRIT receives the image, the tail of the textual prefix, and the candidates \(\mathcal C_i\). It should be noted that GRIT does not receive the original question at this stage, as we only aim to mitigate possible hallucinations occurring in the current step. The model is required to resolve the content of this step alone, rather than the complete question. We obtain (i) a choice \(w^\star\in\mathcal C_i\), (ii) a single human-readable evidence sentence \(\mathcal{E}_i\) from its answer. We then force \(w^\star\) at step \(i\) and append the sentence to the evidence pool:
\begin{equation}
E_{i+1}\;\leftarrow\;E_i\cup\{\mathcal{E}_i\}.
\end{equation}

\textbf{Representation.}
Crucially, only the text participates in scoring via \cref{eq5,eq6,eq7}, and coordinates are stored as annotations for interpretability and for binding the sentence to concrete regions but never participate in scoring steps. This design is deliberate. Injecting raw crops back into the context, such as in zoom-in pipelines, repeatedly couples pixel processing with the entire reasoning chain and typically requires additional supervision or preference optimization for the model to learn when and where to zoom. In contrast, textual evidence is compact, semantic, and model-native: it lives in the same token space as the decoder, can be scored without revisiting pixels, and naturally composes with prior sentences. This keeps the intervention lightweight while saving a verifiable trail.

\textbf{Initialization and growth.}
We initialize \(E_0\) with a global description \(d_{\text{global}}\) to provide broad coverage but not to act as the sole evidence source. Thereafter the evidence pool grows only on demand. This maintains modest computation, preserves determinism in non-ambiguous regions, and lets evidence accumulate precisely where it is most useful for subsequent decisions.

\textbf{Semantics and micro-views.}
Although textual, each sentence implicitly refers to one or multiple related subviews of the image, especially when GRIT generates the answer with coordinates. Because every sentence is generated to resolve a specific local choice, the evidence pool accumulates a set of semantically linked micro-observations that remain logically connected to the evolving chain of thought. \cref{eq6,eq7} convert these micro-observations into reusable probability mass, allowing later steps to benefit from earlier visual disambiguation without re-encoding crops. Therefore ECRD is a bridge that allows base model to interact with a collection of subviews organized by reasoning needs, rather than with isolated cropped regions, thereby enabling reuse across the whole sequence.

%% file: sec/4_experiment.tex
\begin{table*}[t]
    \centering\small
    \setlength{\tabcolsep}{3pt}
    \newcommand{\hh}[2]{\makecell[c]{#1\\#2}}
    \definecolor{gain}{RGB}{37,196,0}
    \newcommand{\gsub}[1]{\raisebox{-0.65ex}{\scriptsize\textcolor{gain}{$\uparrow$\,#1}}}
    \newcommand{\zsub}{\raisebox{-0.65ex}{\scriptsize\textcolor{gray}{0.0}}}
    \resizebox{1.0\textwidth}{!}{
    \begin{tabular}{l cccccccccccc}
    \toprule
    & & \multicolumn{5}{c}{Perception} & \multicolumn{5}{c}{Reasoning} \\
    \cmidrule(lr){3-7}
    \cmidrule(lr){8-12}
    & \textbf{Overall} &
      \hh{Attr.}{} & \hh{Mater.}{} & \hh{Phys.}{} & \hh{ObjRet.}{} & \hh{OCR}{} &
      \hh{Persp.}{} & \hh{Order.}{} & \hh{Cont.\&Oc.}{} & \hh{Contain.}{} & \hh{Compar.}{} \\
    \midrule
    \multicolumn{12}{c}{\textbf{Private Models}} \\
    \midrule
    Gemini-2.5-Flash-0520~\cite{gemini-2.5-flash} & 45.9 & 48.3 & 53.9 & 69.6 & 68.8 & 75.0 & 15.3 & 19.3 & 56.1 & 72.4 & 43.2 \\
    GPT-4o-1120~\cite{gpt4o} & 46.9 & 51.7 & 61.5 & 65.2 & 43.8 & 69.1 & 18.8 & 38.6 & 48.8 & 72.4 & 43.2 \\
    Gemini-2.5-Pro-0605~\cite{gemini-2.5-pro} & 54.1 & 51.7 & 61.5 & 56.5 & 75.0 & 83.8 & 20.0 & 36.8 & 65.9 & 86.2 & 54.6 \\
    o3-0416~\cite{o3} & 54.8 & 69.0 & 69.2 & 65.2 & 68.8 & 79.4 & 22.4 & 38.6 & 61.0 & 86.2 & 50.0 \\ 
    \midrule
    \multicolumn{12}{c}{\textbf{Open-source General Models}} \\
    \midrule
    Qwen2.5-VL-7B~\cite{bai2025qwen25vl} & 37.0 & 55.2 & 53.8 & 56.5 & 62.5 & 27.9 & 20.0 & 35.1 & 39.0 & 44.8 & 43.2 \\
    \rowcolor{gray!15}
    \textbf{\quad\quad\quad\quad\quad\quad + ECRD} &
      47.9\gsub{10.9} & \textbf{72.4}\gsub{17.2} & 53.8\zsub & 73.9\gsub{17.4} & 62.5\zsub & 54.4\gsub{26.5} &
      20.0\zsub & 35.1\zsub & 56.1\gsub{17.1} & 75.9\gsub{31.1} & 45.5\gsub{2.3} \\
    Qwen2.5-VL-32B~\cite{bai2025qwen25vl} & 42.5 & 51.7 & 53.8 & 69.6 & 62.5 & 54.4 & 16.5 & 33.3 & 46.3 & 62.1 & 38.6 \\
    \rowcolor{gray!15}
    \textbf{\quad\quad\quad\quad\quad\quad + ECRD} &
      48.6\gsub{6.1} & 62.1\gsub{10.4} & 53.8\zsub & 73.9\gsub{4.3} & 62.5\zsub & 60.3\gsub{5.9} &
      23.5\gsub{7.0} & 35.1\gsub{1.8} & 61.0\gsub{14.7} & 65.5\gsub{3.4} & 45.5\gsub{6.9} \\
    Qwen2.5-VL-72B~\cite{bai2025qwen25vl} & 42.2 & 65.5 & \textbf{69.2} & 56.5 & 56.3 & 48.5 & 11.8 & 33.3 & 51.2 & 72.4 & 38.6 \\
    \rowcolor{gray!15}
    \textbf{\quad\quad\quad\quad\quad\quad + ECRD} &
      49.9\gsub{7.7} & 65.5\zsub & \textbf{69.2}\zsub & 65.2\gsub{8.7} & 62.5\gsub{6.2} & \textbf{69.1}\gsub{20.6} &
      20.0\gsub{8.2} & \textbf{36.8}\gsub{3.5} & 58.5\gsub{7.3} & 75.9\gsub{3.5} & 40.9\gsub{2.3} \\
    LLaVA-OneVision-7B~\cite{li2024llavaov} & 37.3 & 55.2 & 53.8 & 56.5 & 50.0 & 32.4 & 21.2 & 22.8 & 41.5 & 72.4 & 36.4 \\
    \rowcolor{gray!15}
    \textbf{\quad\quad\quad\quad\quad\quad + ECRD} &
      43.5\gsub{6.2} & 55.2\zsub & 61.5\gsub{7.7} & 60.9\gsub{4.4} & 56.3\gsub{6.3} & 47.1\gsub{14.7} &
      23.5\gsub{2.3} & 28.0\gsub{5.2} & 53.7\gsub{12.2} & 72.4\zsub & 40.9\gsub{4.5} \\
    LLaVA-OneVision-72B~\cite{li2024llavaov} & 40.5 & 62.1 & 53.8 & 65.2 & 62.5 & 36.8 & 12.9 & 28.1 & 53.7 & 65.5 & 47.7 \\
    \rowcolor{gray!15}
    \textbf{\quad\quad\quad\quad\quad\quad + ECRD} &
      46.9\gsub{6.4} & 62.1\zsub & 61.5\gsub{7.7} & 69.6\gsub{4.4} & 68.8\gsub{6.3} & 51.5\gsub{14.7} &
      18.8\gsub{5.9} & 29.8\gsub{1.7} & 61.0\gsub{7.3} & 72.4\gsub{6.9} & \textbf{52.3}\gsub{4.6} \\
    InternVL3-8B~\cite{zhu2025internvl3} & 38.8 & 51.7 & \textbf{69.2} & 56.5 & 56.3 & 33.7 & 21.2 & 24.6 & 39.0 & 72.4 & 43.2 \\
    \rowcolor{gray!15}
    \textbf{\quad\quad\quad\quad\quad\quad + ECRD} &
      45.2\gsub{6.4} & 51.7\zsub & \textbf{69.2}\zsub & 69.6\gsub{13.1} & 62.5\gsub{6.2} & 50.0\gsub{16.3} &
      \textbf{24.7}\gsub{3.5} & 29.8\gsub{5.2} & 43.9\gsub{4.9} & 72.4\zsub & 50.0\gsub{6.8} \\
    InternVL3-38B~\cite{zhu2025internvl3} & 42.0 & 51.7 & 61.5 & 52.2 & 68.8 & 51.5 & 12.9 & 33.3 & 56.1 & 65.5 & 38.6 \\
    \rowcolor{gray!15}
    \textbf{\quad\quad\quad\quad\quad\quad + ECRD} &
      48.4\gsub{6.4} & 51.7\zsub & \textbf{69.2}\gsub{7.7} & 65.2\gsub{13.0} & \textbf{75.0}\gsub{6.2} & 58.8\gsub{7.3} &
      22.4\gsub{9.5} & \textbf{36.8}\gsub{3.5} & 56.1\zsub & 69.0\gsub{3.5} & 50.0\gsub{11.4} \\
    InternVL3-78B~\cite{zhu2025internvl3} & 46.4 & 62.1 & 61.5 & 52.2 & 68.8 & 52.9 & 16.5 & 33.3 & 61.0 & \textbf{86.2} & 45.5 \\
    \rowcolor{gray!15}
    \textbf{\quad\quad\quad\quad\quad\quad + ECRD} &
      \textbf{50.9}\gsub{4.5} & 65.5\gsub{3.4} & \textbf{69.2}\gsub{7.7} & 65.2\gsub{13.0} & \textbf{75.0}\gsub{6.2} & 60.3\gsub{7.4} &
      21.2\gsub{4.7} & 35.1\gsub{1.8} & \textbf{63.4}\gsub{2.4} & \textbf{86.2}\zsub & 47.7\gsub{2.2} \\
    \midrule
    \multicolumn{12}{c}{\textbf{Open-source Visual Grounded Reasoning Models}} \\
    \midrule
    DeepEyes-7B~\cite{zheng2025deepeyes} & 37.5 & 62.1 & 53.8 & 65.2 & 68.8 & 51.5 & 11.8 & 24.6 & 36.6 & 51.7 & 47.7 \\
    Pixel-Reasoner-7B~\cite{su2025pixel} & 39.0 & 58.6 & 61.5 & 65.2 & 50.0 & 48.5 & 14.1 & 31.6 & 39.0 & 44.8 & 40.9 \\
    TreeVGR-7B~\cite{wang2025traceable} & 50.4 & 65.5 & 53.8 & \textbf{82.6} & 68.8 & 63.3 & 22.4 & \textbf{36.8} & 61.0 & 69.0 & 45.5 \\
    \bottomrule
    \end{tabular}}
    \vspace{-4pt}
    \caption{Results of different models on TreeBench. Best performances for open-source models are highlighted in \textbf{bold}. ECRD consistently improves open-source general models across architectures and scales, confirming training-free plug-and-play applicability.}
    \vspace{-14pt}
    \label{table1}
\end{table*}

\section{Experiment}
\label{sec:experiment}

\subsection{Setups}

We evaluate ECRD across benchmarks grouped by the capabilities they probe: (i) visual grounded reasoning under long chains, (ii) reasoning–perception balance and hallucination control, and (iii) broad multimodal competence. In the experiments reported in \cref{table1,table2,table3,tab:treebench-ablation,tab:treebench-trainingfree}, we set the uncertainty threshold $\delta=0.08$. No additional training are used.

\subsection{Benchmarks and base models}

\textbf{Benchmarks.}
We evaluate on three groups of benchmarks. TreeBench~\cite{wang2025traceable} probes \enquote{thinking with images} by separating Perception (identify, localize, read, attribute) from Reasoning (operate over aggregated visual evidence such as occlusion, containment, ordering, perspective); the metric is answer accuracy. RH-Bench~\cite{liu2025thinkingseeingassessingamplified} reports Reason and Perception scores and RH-AUC~\cite{liu2025thinkingseeingassessingamplified}, which summarizes the trade-off between reasoning length and hallucination (higher is better). For general multimodal competence, we use V*Bench~\cite{wu2024v}, MathVista~\cite{lu2023mathvista}, ChartQA~\cite{masry2022chartqa}, OCRBench~\cite{liu2024ocrbench}, and HallusionBench~\cite{guan2024hallusionbench}, evaluated by accuracy.

\textbf{Base models.}
Unless otherwise specified, the visual decider is GRIT-3B~\cite{fan2025grit}, which is built upon Qwen2.5-VL-3B~\cite{bai2025qwen25vl} and optimized for visual grounding. To assess plug-and-play generality and scale robustness, we also attach ECRD without any finetuning to three open-source general model families: LLaVA-OneVision~\cite{li2024llavaov} (7B and 72B), Qwen2.5-VL~\cite{bai2025qwen25vl} (7B, 32B and 72B), and InternVL3~\cite{zhu2025internvl3} (8B, 38B and 78B). All checkpoints and inference recipes follow the authors’ official releases. ECRD modifies only the decoding procedure at test time and leaves the underlying encoders or decoders frozen. For reference lines on TreeBench, we also report private models GPT-4o~\cite{gpt4o} and o3~\cite{o3} as well as Gemini-2.5-Flash~\cite{gemini-2.5-flash} and Gemini-2.5-Pro~\cite{gemini-2.5-pro}, and include recent RL-based visually-grounded reasoning systems DeepEyes~\cite{zheng2025deepeyes}, Pixel-Reasoner~\cite{su2025pixel}, and TreeVGR-7B~\cite{wang2025traceable} for comparison. The paper that presents the TreeVGR-7B~\cite{wang2025traceable} method also introduces TreeBench.

\subsection{Main results and analysis}

As demonstrated in \cref{table1}, ECRD delivers consistent gains across all open-source general backbones and scales on TreeBench, showing that the method is truly plug-and-play rather than model-specific. On Qwen2.5-VL-7B the overall accuracy rises from 37.0\% to 47.9\%, and similar improvements appear on LLaVA-OneVision-7B and InternVL3-8B, as well as on larger models such as Qwen2.5-VL-32B/72B, LLaVA-OneVision-72B, and InternVL3-38B/78B, where the absolute lifts are typically in the +4-8 point band. The sub-category pattern matches our design: competencies that rely on checkable visual facts such as OCR, Physical State and Comparison benefit the most. Compared to RLHF-based visual-grounded reasoning models, ECRD on Qwen2.5-VL-7B surpasses DeepEyes-7B and Pixel-Reasoner-7B and approaches TreeVGR-7B, yet it requires no additional training, curated traces, or reinforcement optimization. It also exceeds several strong closed models such as Gemini-2.5-Flash and GPT-4o while still trailing Gemini-2.5-Pro and o3.

To further position ECRD against established training-free baselines beyond RL-based systems, we compare it on TreeBench using the same base model (Qwen2.5-VL-7B) with: inference-time correction (Woodpecker~\cite{yin2024woodpecker}), programmatic reasoning (ViperGPT~\cite{suris2023vipergpt}), and visual prompting (ControlMLLM~\cite{wu2024controlmllm}). We also evaluate standard decoding alternatives, including beam search, self-consistency, and diverse sampling. As shown in \cref{tab:treebench-trainingfree}, all these baselines underperform ECRD, indicating a significantly better performance of ECRD among plug-and-play approaches.

\begin{table}[t]
  \centering
  \setlength{\tabcolsep}{3.2pt}
  \small
  \resizebox{\columnwidth}{!}{
  \begin{tabular}{lcccccccc}
    \toprule
    \textbf{Method} &
    Base &
    Woodpecker &
    ViperGPT &
    ControlMLLM &
    Beam &
    Self-cons. &
    Diverse &
    \textbf{ECRD} \\
    \midrule
    \textbf{Overall} &
    37.0 &
    36.3 &
    38.3 &
    40.5 &
    38.8 &
    39.0 &
    38.3 &
    \textbf{47.9} \\
    \bottomrule
  \end{tabular}}
  \vspace{-6pt}
  \caption{Comparison with training-free baselines and decoding alternatives on TreeBench. Best result are highlighted in \textbf{bold}. ECRD performs markedly better among all plug-and-play settings.}
  \vspace{-16pt}
  \label{tab:treebench-trainingfree}
\end{table}

\begin{table*}[t]
    \centering\small
    \setlength{\tabcolsep}{3pt}
    \newcommand{\hh}[2]{\makecell[c]{#1\\#2}}
    \resizebox{1.0\textwidth}{!}{
    \begin{tabular}{l cccccccccccc}
    \toprule
    & & \multicolumn{5}{c}{Perception} & \multicolumn{5}{c}{Reasoning} \\
    \cmidrule(lr){3-7}\cmidrule(lr){8-12}
    & \textbf{Overall} &
      \hh{Attr.}{} & \hh{Mater.}{} & \hh{Phys.}{} & \hh{ObjRet.}{} & \hh{OCR}{} &
      \hh{Persp.}{} & \hh{Order.}{} & \hh{Cont.\&Oc.}{} & \hh{Contain.}{} & \hh{Compar.}{} \\
    \midrule
    GRIT-3B~\cite{fan2025grit} & 30.1 & 31.0 & 23.1 & 56.5 & 25.0 & 39.7 & \textbf{21.2} & 17.5 & 36.6 & 48.3 & 20.5 \\
    Qwen2.5-VL-7B~\cite{bai2025qwen25vl} & 37.0 & 55.2 & \textbf{53.8} & 56.5 & \textbf{62.5} & 27.9 & 20.0 & \textbf{35.1} & 39.0 & 44.8 & 43.2 \\
    \textcolor{gray}{Qwen2.5-VL-7B + VDGD} & \textcolor{gray}{39.5} & \textcolor{gray}{58.6} & \textcolor{gray}{\textbf{53.8}} & \textcolor{gray}{52.2} & \textcolor{gray}{\textbf{62.5}} & \textcolor{gray}{50.0} & \textcolor{gray}{17.6} & \textcolor{gray}{29.8} & \textcolor{gray}{39.0} & \textcolor{gray}{48.3} & \textcolor{gray}{40.9} \\
    {\makecell[l]{\textcolor{gray}{\textbf{Qwen2.5-VL-7B + supervisor}}}}
    & \textcolor{gray}{40.7} & \textcolor{gray}{58.6} & \textcolor{gray}{\textbf{53.8}} & \textcolor{gray}{56.5} & \textcolor{gray}{\textbf{62.5}} & \textcolor{gray}{51.5} & \textcolor{gray}{18.8} & \textcolor{gray}{31.6} & \textcolor{gray}{41.5} & \textcolor{gray}{44.8} & \textcolor{gray}{43.2} \\
    {\makecell[l]{\textcolor{gray}{\textbf{Qwen2.5-VL-7B + ECRD}} \\ \textcolor{gray}{\textbf{(Qwen2.5-VL-3B as visual decider)}}}}
    & \textcolor{gray}{43.7} & \textcolor{gray}{62.1} & \textcolor{gray}{\textbf{53.8}} & \textcolor{gray}{69.6} & \textcolor{gray}{\textbf{62.5}} & \textcolor{gray}{52.9} & \textcolor{gray}{20.0} & \textcolor{gray}{31.6} & \textcolor{gray}{43.9} & \textcolor{gray}{62.1} & \textcolor{gray}{43.2} \\
    \rowcolor{gray!15}
    \textbf{Qwen2.5-VL-7B + ECRD} & \textbf{47.9} & \textbf{72.4} & \textbf{53.8} & \textbf{73.9} & \textbf{62.5} & \textbf{54.4} & 20.0 & \textbf{35.1} & \textbf{56.1} & \textbf{75.9} & \textbf{45.5} \\
    \bottomrule
    \end{tabular}}
    \vspace{-4pt}
    \caption{TreeBench ablations on Qwen2.5-VL-7B. Best performances are highlighted in \textbf{bold}. The supervisor provides a stable boost and the visual decider adds the remaining lift; both components are necessary and complementary.}
    \vspace{-8pt}
    \label{tab:treebench-ablation}
\end{table*}

\begin{table}[t]
  \centering
  \setlength{\tabcolsep}{4pt}
  \begin{tabular}{lccc}
    \toprule
    \multicolumn{1}{c}{\textbf{Model}} & \multicolumn{3}{c}{\textbf{RH-Bench}} \\
    \cmidrule(lr){2-4}
    & Reas. & Perc. & RH-AUC \\
    \midrule
    Qwen2.5-VL-7B~\cite{bai2025qwen25vl} & 39.6 & 50.2 & 0.51 \\
    {\makecell[l]{\textcolor{gray}{\textbf{Qwen2.5-VL-7B + supervisor}}}}
    & \textcolor{gray}{42.0} & \textcolor{gray}{53.3} & \textcolor{gray}{0.54} \\
    \rowcolor{gray!15}
    \textbf{Qwen2.5-VL-7B + ECRD} & \textbf{46.4} & \textbf{57.1} & \textbf{0.58} \\
    \bottomrule
  \end{tabular}
  \vspace{-4pt}
  \caption{Results on RH-Bench. Reas.\ indicates \textit{Reasoning}, and Perc.\ indicates \textit{Perception}. Best performances are highlighted in \textbf{bold}. ECRD improves both Reasoning and Perception while increasing RH-AUC, indicating a better balance between reasoning and hallucination over longer chains.}
  \vspace{-8pt}
  \label{table2}
\end{table}

\begin{table*}[t]
  \centering
  \setlength{\tabcolsep}{5pt}
  \begin{tabular}{lccccccc}
    \toprule
    \multicolumn{1}{c}{\textbf{Model}} &
    \multicolumn{3}{c}{\textbf{V* Bench}} &
    \textbf{MathVista} & \textbf{ChartQA} & \textbf{OCRBench} & \textbf{HallusionBench} \\
    \cmidrule(lr){2-4}
    & Attr & Spatial & Overall &  &  &  &  \\
    \midrule
    LLaVA-OneVision-7B~\cite{li2024llavaov}        & 73.0 & 60.5 & 68.1 & 63.2 & 80.0 & 62.2 & 55.1 \\
    \rowcolor{gray!15}
    \textbf{LLaVA-OneVision-7B + ECRD} & \textbf{74.8} & \textbf{65.8} & \textbf{71.2} & \textbf{67.9} & \textbf{86.3} & \textbf{73.8} & \textbf{63.7} \\
    Qwen2.5-VL-7B~\cite{bai2025qwen25vl}        & 73.9 & 67.1 & 71.2 & 68.2 & 84.4 & 82.3 & 61.3 \\
    {\makecell[l]{\textcolor{gray}{\textbf{Qwen2.5-VL-7B + supervisor}}}}
    & \textcolor{gray}{73.9} & \textcolor{gray}{71.1} & \textcolor{gray}{72.8} & \textcolor{gray}{69.8} & \textcolor{gray}{86.8} & \textcolor{gray}{85.7} & \textcolor{gray}{67.5} \\
    \rowcolor{gray!15}
    \textbf{Qwen2.5-VL-7B + ECRD} & \textbf{74.8} & \textbf{75.0} & \textbf{74.9} & \textbf{72.3} & \textbf{88.3} & \textbf{90.7} & \textbf{72.5} \\
    \bottomrule
  \end{tabular}
  \vspace{-4pt}
  \caption{Results on five general multimodal benchmarks. Better performances are highlighted in \textbf{bold}. ECRD brings consistent gains across diverse tasks on both Qwen2.5-VL-7B and LLaVA-OneVision-7B, demonstrating backbone-agnostic, task-general effectiveness.}
  \vspace{-8pt}
  \label{table3}
\end{table*}

\cref{table2} shows that ECRD improves Reasoning from 39.6\% to 46.4\% and Perception from 50.2\% to 57.1\%, lifting RH-AUC from 0.51 to 0.58 on RH-Bench . A higher RH-AUC value indicates that as the chain grows longer, accuracy remains at a higher level and the balance between reasoning and hallucination is better maintained.

Finally, across five general multimodal benchmarks in \cref{table3}, ECRD brings broad, task-general gains on both Qwen2.5-VL-7B and LLaVA-OneVision-7B: V*Bench overall rises by several points, MathVista and ChartQA see steady improvements, OCRBench jumps by around +8–12 points, and HallusionBench improves by roughly +8–11 points, reflecting fewer visually induced slips that would otherwise propagate through the chain.

\subsection{Ablation study}

The ablations on Qwen2.5-VL-7B in \cref{tab:treebench-ablation} isolate where the lift comes from. First, GRIT-3B alone yields the lowest TreeBench accuracy among all rows, yet ECRD, invoking GRIT-3B only at uncertain steps, reaches 47.9\% (+17.8 over GRIT-3B and +10.9 over the 7B base), which rules out the hypothesis that gains are driven by a stronger perception model and points instead to the decoding design, specifically ECRD’s ability to fully leverage the small model exactly where it matters. Second, replacing VDGD’s prefix-wise minimum with our mean-over-prefix evidence scoring already helps without any decide, confirming that averaging stabilizes token selection. Third, adding a grounded decider matters: using Qwen2.5-VL-3B, a generic VLM, as the decider improves further, and swapping in the grounding-oriented GRIT-3B yields the full ECRD gains, with the largest margins on visually consequential categories. The same staged pattern appears beyond TreeBench: as shown in \cref{table2,table3}, on RH-Bench and on the five general multimodal benchmarks, the variant without the decider consistently lies between the base and the full method, showing that both parts are necessary: the supervisor provides a robust, always-on stabilization under uncertainty, and the visual decider supplies sparse but decisive micro-observations only when ambiguity persists, converting the remaining hard steps and propagating benefits through rest of chain.

\begin{figure*}[t]
\centering
\captionsetup{skip=-6pt}
\includegraphics[width=2.1\columnwidth]{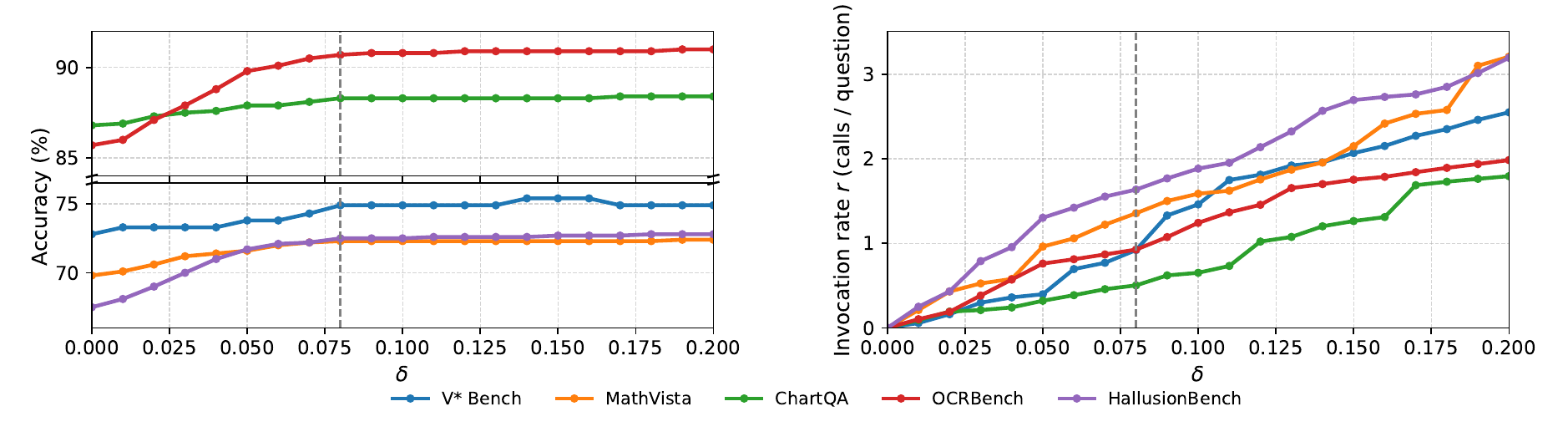}
\caption{Analysis of the uncertainty threshold $\delta$: accuracy as a function of $\delta$ across five benchmarks, and the average visual decider invocation rate (calls per question) as $\delta$ varies. The gray dashed line marks $\delta=0.08$.}
\vspace{-11pt}
\label{fig3}
\end{figure*}

\begin{table}[t]
\centering
\setlength{\tabcolsep}{3pt}
\resizebox{\columnwidth}{!}{
\begin{tabular}{lccccc}
\toprule
Benchmark & V* Bench & MathVista & ChartQA & OCRBench & HallusionBench \\
\midrule
$t_0$ & 8.98 & 12.92 & 9.76 & 3.24 & 11.67 \\
$l_0$ & 1.32 & 1.46 & 1.30 & 1.12 & 1.43 \\
\bottomrule
\end{tabular}
}
\vspace{-4pt}
\caption{Average time per question at $\delta=0$ ($t_0$, seconds per question) and global average latency of a single visual decider call ($l_0$, seconds). Note that $l_0$ is computed as an overall mean across all benchmarks and $\delta$ settings, and is not restricted to the $\delta=0$ condition. All tests are conducted on a single H20-NVLink GPU.}
\vspace{-12pt}
\label{table4}
\end{table}

\begin{figure*}[t]
\centering
\captionsetup{skip=4pt}
\includegraphics[width=2.1\columnwidth]{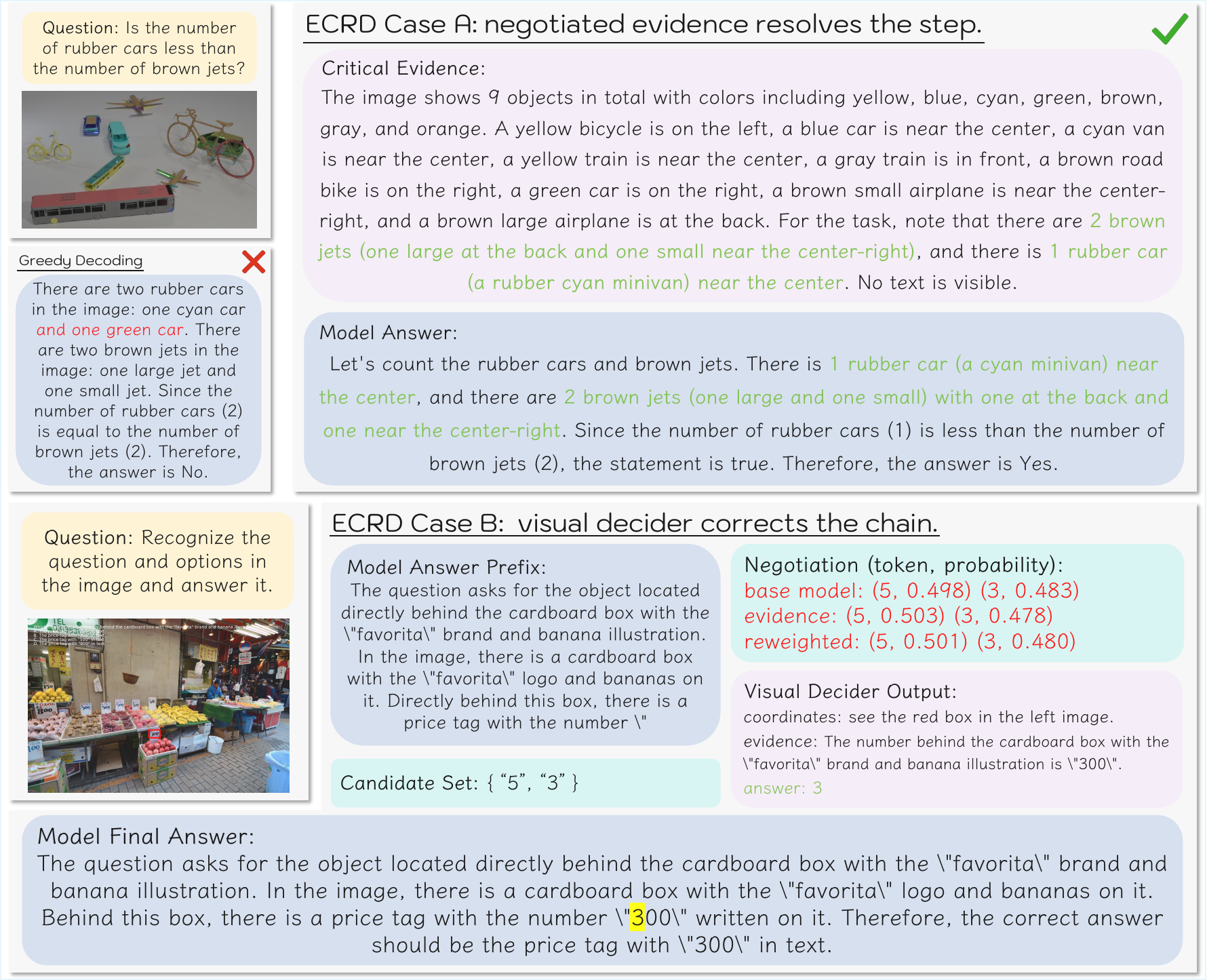}
\caption{Two typical application cases of ECRD.}
\vspace{-11pt}
\label{fig4}
\end{figure*}

\begin{figure}[t]
\centering
\captionsetup{skip=4pt}
\includegraphics[width=0.85\columnwidth]{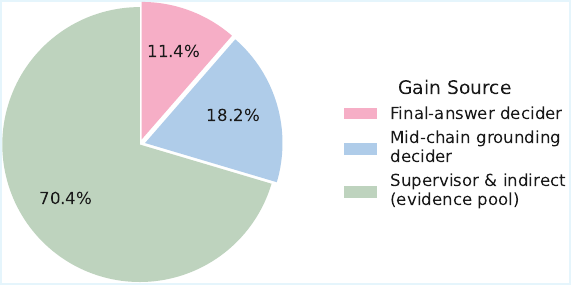}
\caption{Breakdown of ECRD’s gain on TreeBench based on Qwen2.5-VL-7B.}
\vspace{-16pt}
\label{fig5}
\end{figure}

\subsection{Efficiency analysis and uncertainty threshold}

ECRD adds two sources of computation on top of the base decoder: (i) evidence scoring over the knee-selected candidate set and (ii) decider calls triggered when uncertainty persists. The former is lightweight. At step \(t\), after knee truncation produces \(\mathcal C_t =\{w_1,\ldots,w_k\}\), scoring requires a pass over the evidence pool \(E_t\) to form the KL-style preference \(s(w_i|E_t)\). Because each evidence contributes precomputed log-likelihoods on a separate and cache-disabled backend (stored as \(\text{FP16}\) on CPU), the inference-time computation is \(O(k|E_t|)\) with negligible GPU pressure. In practice \(k\) is single-digit and \(|E_t|\) grows slowly, so this path contributes only a small fraction of total latency.

The heavier component is the visual decider. Let $r$ denote the average number of decider calls per question. As \cref{fig3} shows, increasing the uncertainty threshold $\delta$ monotonically raises $r$ (right panel), while accuracy (left panel) exhibits a consistent elbow: a rapid lift for small $\delta$, followed by saturation. The knee for all five benchmarks concentrates around the gray dashed line ($\delta \approx 0.08$), which is precisely where the negotiated distribution most often flags genuinely ambiguous, visually charged steps. Benchmarks whose items hinge on precise perceptual grounding such as OCRBench and HallusionBench benefit early and strongly, their curves rise steeply with a small number of calls, whereas tasks that already align well with text-only priors like ChartQA show milder, earlier saturation. MathVista and V* Bench lie in between: accuracy improves steadily as a handful of targeted micro-observations are injected, then flattens when further calls add little new information. Importantly, beyond the knee, the invocation of visual decider grows faster than accuracy, and some curves even level off or gently fluctuate, reflecting diminishing returns once the few decisive ambiguities have been resolved.

The cost trend aligns with a simple latency model grounded in \cref{table4}. Let $t_0$ be the base time per question at $\delta=0$ (no calls) and $l_0$ be the average marginal latency of a single call. Then the end-to-end time obeys
\begin{equation}
T(\delta) \approx t_0 + l_0\, r(\delta).
\end{equation}
with $t_0$ varying by benchmark and $l_0$ staying in a narrow band (single-second scale) across settings. Combined with \cref{fig3}, this explains the observed cost–accuracy balance: near $\delta \approx 0.08$, accuracy captures most of the attainable gain while $r$ remains in the low single-digit regime, yielding modest overhead relative to $t_0$, whereas pushing $\delta$ higher mainly increases $r$ (and thus $T$) with little additional accuracy. 
Hence we adopt $\delta=0.08$ as a default: it lies at the elbow where added calls cease to be cost-effective, yet the model already recovers most of the gains in \cref{table1,table2,table3,tab:treebench-ablation,tab:treebench-trainingfree}.

\subsection{Qualitative analysis: how ECRD works}

As shown in \cref{fig2,fig4}, we present three representative cases that correct typical failure modes, corresponding to ECRD’s two components—the supervisor and the visual decider: (i) negotiated reweighting with textual evidence and (ii) on-demand visual arbitration when uncertainty persists.

\textbf{Case A: Negotiated evidence resolves the step.}
As shown in \cref{fig4}, the question asks whether rubber cars are fewer than brown jets. At critical steps, ECRD does not trigger the decider because of the large reweighted gap, instead, it reweights the candidates so that tokens consistent with evidence are favored, yielding the correct answer \enquote{Yes}.

\textbf{Case B.1: Visual decider supplies mid-chain grounding.}
As shown in \cref{fig2} (full examples are provided in supplementary material), at the key step the candidate set is \{ “blue”, “red” \}. The base slightly prefers “red”, while the evidence-induced distribution prefers “blue”; the negotiated gap remains small, so the trigger fires. This choice is pivotal for the rest of the chain: if the model commits to “red”, the subsequent localization and description would drift toward the red garment and propagate errors. The visual decider reads the image with the current prefix and returns the grounding sentence: \enquote{The first dress from the right-hand side is blue, partially hidden by the tree.} ECRD forces “blue” for the current step and adds this sentence to the pool. Later tokens then follow this micro-observation to correctly describe all three dresses and reach the right final color judgment for the queried position.

\textbf{Case B.2: Visual decider supplies the final answer.}
As shown in \cref{fig4}, here the candidate set at the decisive step is \{ “5”, “3” \}. The base leans toward “5”, the evidence-induced distribution leans toward “3”, and the small negotiated gap triggers the decider. The visual decider localizes the tag behind the box and returns: \enquote{The number behind the cardboard box with the \enquote{favorita} brand and banana illustration is \enquote{300}.} ECRD commits token “3” and inserts the sentence into the pool; at the next two steps the supervisor prefers tokens consistent with this evidence, selecting “0” then “0” without additional calls. The chain thus outputs the correct answer \enquote{300}.

\textbf{Where the gains come from.}
Based on a detailed statistical analysis of the experimental results, we find that on Qwen2.5-VL-7B, ECRD yields a +10.9-point overall improvement on TreeBench. As shown in \cref{fig5}, within this lift, cases where the visual decider directly outputs the final answer account for 11.4\% of the gain, while cases where the decider injects a mid-chain visual grounding that unlocks the rest of the reasoning account for 18.2\%. The remaining improvement primarily comes from the supervisor’s negotiated reweighting and the indirect benefits of an expanding evidence pool, which stabilizes later token choices even when no further decider calls are needed.

%% file: sec/5_conclusion.tex
\section{Conclusion}
\label{sec:conclusion}

We presented ECRD, a decoding-time, training-free, plug-and-play framework for visually grounded reasoning.
By reconciling base probabilities with an evidence-guided distribution, ECRD preserves model confidence, invokes a lightweight visual decider only when necessary, and propagates visual grounding through reasoning chain.
Experiments show consistent gains across diverse benchmarks.